\documentclass[sigconf, natbib]{acmart}

\usepackage{makecell}
\usepackage{multirow}
\usepackage{subcaption}
\usepackage{graphicx}
\usepackage{tabularx}
\usepackage{makecell}
\usepackage{caption}
\usepackage{url}
\usepackage{enumitem}
\usepackage{array}
\usepackage{booktabs}
\usepackage{amsmath}
\usepackage{pifont}
\usepackage{mdframed}
\usepackage{pifont} 
\usepackage{xspace}
\usepackage{xcolor}
\usepackage{algorithm}
\usepackage[noend]{algpseudocode}
\usepackage{listings}
\usepackage{enumitem}

\definecolor{darkgreen}{RGB}{0, 100, 0}
\newcommand{\mygreen}[1]{\textcolor{darkgreen}{#1}}
\newcommand{\toolname}[1]{\texttt{\textcolor{darkgreen}{#1}}}

\newcommand{\soften}[1]{\textcolor{gray}{\small #1}}

\newcommand{\policy}{$\mathit{P}$}
\newcommand{\format}{$\mathit{D}_{int}$}
\newcommand{\augment}{$\mathit{D}_{aug}$}

\newcommand{\policyllm}{$\mathit{P}_{\mathit{LLM}}$}

\newcommand{\method}{\textsc{OMuleT}\xspace}
\newcommand{\methodfull}{\underline{O}rchestrating \underline{Mul}tipl\underline{e} \underline{T}ools\xspace}

\newcolumntype{Z}{>{\raggedright\arraybackslash}X} 
\newcolumntype{S}{>{\hsize=.4\hsize}Z} 
\newcolumntype{T}{>{\hsize=.8\hsize}Z}

\newcommand{\boldheading}[1]{%
    \vspace{0.5em} 
    \noindent\textbf{#1}\hspace{0.1em} 
}

\AtBeginDocument{%
  \providecommand\BibTeX{{%
    \normalfont B\kern-0.5em{\scshape i\kern-0.25em b}\kern-0.8em\TeX}}}

\setcopyright{acmcopyright}
\copyrightyear{2024}
\acmYear{2024}
\acmDOI{tbd}

\acmConference[Conf]{Conf}{Month DD, YYYY}{City, State}

\begin{document}

\title{OMuleT: Orchestrating Multiple Tools for Practicable Conversational Recommendation}

\author{Se-eun Yoon}
\authornote{This work was done during the author's internship at Roblox.}
\affiliation{%
  \institution{University of California, San Diego}
  \city{La Jolla}
  \state{CA}
  \country{USA}}
\email{seeuny@ucsd.edu}

\author{Xiaokai Wei}
\affiliation{%
  \institution{Roblox}
  \city{San Mateo}
  \state{CA}
  \country{USA}}
\email{xwei@roblox.com}

\author{Yexi Jiang}
\affiliation{%
  \institution{Roblox}
  \city{San Mateo}
  \state{CA}
  \country{USA}}
\email{yjiang@roblox.com}

\author{Rachit Pareek}
\affiliation{%
  \institution{Roblox}
  \city{San Mateo}
  \state{CA}
  \country{USA}}
\email{rpareek@roblox.com}

\author{Frank Ong}
\affiliation{%
  \institution{Roblox}
  \city{San Mateo}
  \state{CA}
  \country{USA}}
\email{fong@roblox.com}

\author{Kevin Gao}
\affiliation{%
  \institution{Roblox}
  \city{San Mateo}
  \state{CA}
  \country{USA}}
\email{kgao@roblox.com}

\author{Julian McAuley}
\affiliation{%
  \institution{University of California, San Diego}
  \city{La Jolla}
  \state{CA}
  \country{USA}}
\email{jmcauley@ucsd.edu}

\author{Michelle Gong}
\affiliation{%
  \institution{Roblox}
  \city{San Mateo}
  \state{CA}
  \country{USA}}
\email{mgong@roblox.com}


\renewcommand{\shortauthors}{Se-eun Yoon, Xiaokai Wei, Yexi Jiang, Rachit Pareek, Frank Ong, Kevin Gao, Julian McAuley, and Michelle Gong}

\begin{abstract}
In this paper, we present a systematic effort to design, evaluate, and implement a realistic conversational recommender system (CRS).
The objective of our system is to allow users to input free-form text to request recommendations, and then receive a list of relevant and diverse items.
While previous work on synthetic queries augments large language models (LLMs) with 1-3 tools, we argue that a more extensive toolbox is necessary to effectively handle real user requests.
As such, we propose a novel approach that equips LLMs with over 10 tools, providing them access to the internal knowledge base and API calls used in production. 
We evaluate our model on a dataset of real users and show that it generates relevant, novel, and diverse recommendations compared to vanilla LLMs.
Furthermore, we conduct ablation studies to demonstrate the effectiveness of using the full range of tools in our toolbox.
We share our designs and lessons learned from deploying the system for internal alpha release.
Our contribution is the addressing of all four key aspects of a practicable CRS: 
(1) real user requests, (2) augmenting LLMs with a wide variety of tools, (3) extensive evaluation, and (4) deployment insights.
\end{abstract}





\maketitle

\begin{figure}[t!]
    \centering
    \vspace{1em}
    \includegraphics[width=0.95\linewidth]{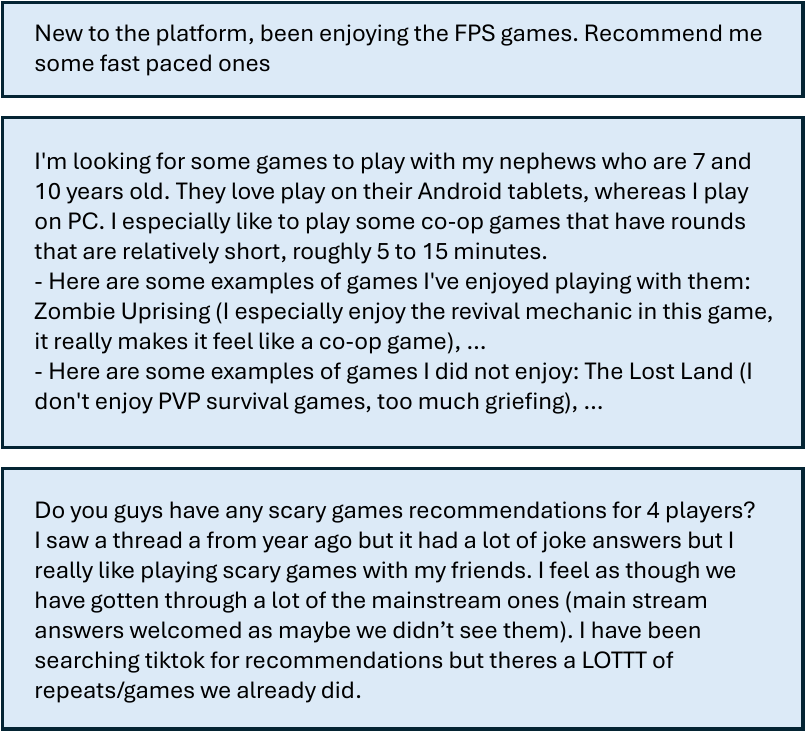}
    \caption{Examples of recommendation requests from users.}
    \Description{}
    \label{fig:figure1}
\end{figure}

\section{Introduction}

Imagine a user who wants to find new games but faces thousands to millions of options.
Since trying out various games can be time-consuming, one may want to get recommendations simply by saying in natural language what they want to play. 
Examples of such user requests are depicted in Figure~\ref{fig:figure1}, where users express their unique needs through diverse expressions.
A conversational recommender system (CRS) that can take in such free-form requests and retrieve the most relevant items would greatly improve user experience in navigating through a vast choice of content.

While there are many works on CRS~\cite{christakopoulou2016towards, sun2018conversational, zhang2018towards, li2018towards, lei2020estimation, zhang2020conversational, wang2022towards}, rarely do we see a system in practice.
Even though large language models (LLMs) have been demonstrated to be effective in conversational movie recommendation~\cite{he2023large, sanner2023large}, LLMs alone cannot be directly applied to many industrial domains.
One limitation of LLMs is their dependence on fixed parameters, which restricts their ability to handle a dynamic pool of items and integrate up-to-date world knowledge without the costly process of fine-tuning.
Furthermore, LLMs exhibit high popularity bias, frequently recommending or addressing the most well-known items~\cite{he2023large}.

This work contributes to practicable CRS research through the following efforts.
First, we collect a dataset of \textit{real user requests} and recommendations.
This distinguishes our work from papers that use synthetic queries generated from traditional user-item interactions~\cite{huang2023recommender, kemper2024retrieval, wang2024recmind}.
Real user requests are more challenging to process than requests synthesized from templates due to their variety, unstructured nature, and subjective language~\cite{he2023large, yoon2024evaluating}.
Second, in order to process such complex requests, we argue that \textit{a much larger number of tools} are required to augment LLMs for recommendations, compared to existing approaches that address synthetic queries with only 1-3 tools.
For example, in real user requests, free-form casual utterances (e.g., using `ptfs' to refer to the game `Pilot Training Flight Simulator') require specialized tools for processing, which is not necessary for synthetic requests that use clearly defined item names.
Another example is handling complex conditions, such as a user who plays games on a PC and wants games to play with 7- and 10-year-old nephews who use tablets, and providing a list of liked and disliked games and reasons (see Figure~\ref{fig:figure1}).
Using just a search API~\cite{friedman2023leveraging} or a lookup API~\cite{li2024incorporating} may be insufficient for handling such conditions; multiple tools are required to address factors such as games popular among age groups, device compatibility, and similar games search.
While a large number of tools may initially seem daunting to implement, our tools are relatively generic (e.g., unlike tools that require reviews~\cite{kemper2024retrieval}) and can be easily constructed from databases and APIs available in many industry settings. 

Third, we propose \method (\methodfull), a framework for augmenting LLMs with diverse tools to meet complex requests.
Our method translates a user's raw utterance into a formatted intent, applies a tool-execution policy, and then augments the results to the LLM generating the recommendations (see Figure~\ref{fig:method}).
This approach not only makes the system transparent and controllable, but it is more effective in performance than methods where an LLM generate its own tool execution policy~\cite{huang2023recommender, wang2024executable}.
Finally, we perform \textit{extensive evaluation} on two LLMs (LLaMA-405B~\cite{meta2024llama} and GPT-4o~\cite{openai2024gpt4}) and 8 metrics covering factuality, relevance, novelty, and diversity.
Our results show that using our framework is more effective than baseline LLMs, and multiple tools are necessary for the best performance.
We implement our model for internal testing and share our insights for deployment.
To the best of our knowledge, we are the first work to address all the following elements that are essential for a practicable CRS:

\begin{itemize}[leftmargin=10pt, label={$\circ$}]
    \item \textbf{Real user requests.} We use real user requests, which are more complex and diverse than queries synthesized from templates.
    \item \textbf{Framework for augmenting LLMs with nontrivial amount of tools.} Complex requests require the use of a wide variety of tools. Our tools are simple and generic, and our framework effectively orchestrates the tools to augment LLMs.
    \item \textbf{Extensive evaluation.} We conduct evaluations and ablation studies on various metrics and baselines to demonstrate the effectiveness of our approach. 
    \item \textbf{Deployment insights.} We share our designs and lessons learned from deploying the system for internal demonstrations, hoping to provide guidance for practitioners.
\end{itemize}

\section{Problem Formulation}
Given a user's recommendation request in free-form natural language, the agent should return a list of $k$ items.
The success of the task is measured by multiple criteria.
First, the items should be \textbf{relevant} to the request; they should be what the user is asking for.
An ideal approach to evaluate relevance is to get direct feedback from the user who made the request. 
However, in the early stages of model development, obtaining feedback for each iteration is impractical. 
Thus, we construct an evaluation data as a proxy for relevance, which we discuss in Section~\ref{subsec:data}.
Another important criterion is that items should be \textbf{novel}, since the goal of recommendation closely tied to discovery~\cite{vargas2011novelty}; we want to avoid recommending highly popular items that often appear on the platform's front page.
Finally, the collection of recommended items across all requests should have high \textbf{coverage}, ensuring a diverse range of recommendations.
This breadth of visibility is especially crucial for the success of a platform that relies on millions of user-generated content.

\begin{figure}[t!]
    \centering
    \vspace{1em}
    \includegraphics[width=0.90\linewidth]{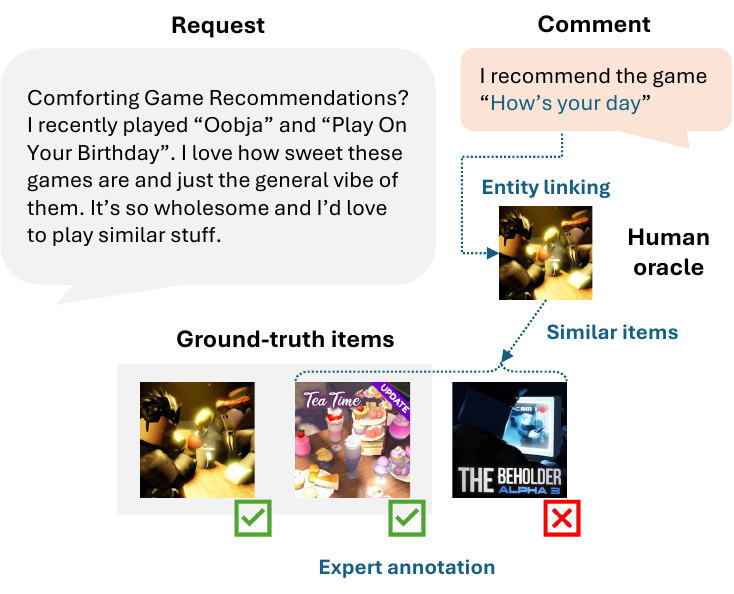}
    \caption{Our dataset collection process.}
    \Description{}
    \label{fig:data}
\end{figure}

\begin{figure*}[t]
    \centering
    \vspace{1em}
    \includegraphics[width=0.95\linewidth]{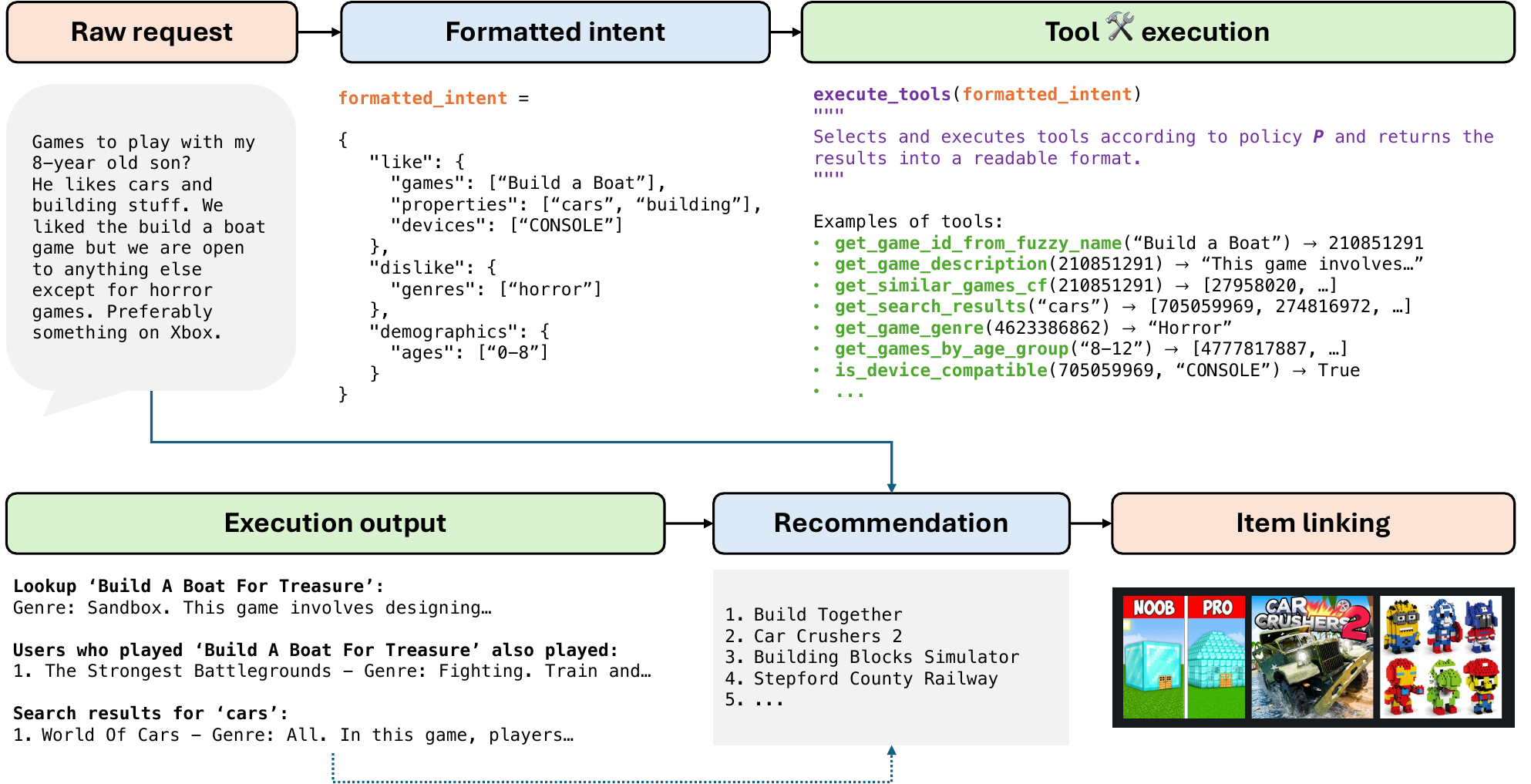}
    \caption{Overview of \method. Orange boxes are in the user interface (a user inputs a raw request and observes recommended items); blue boxes are where LLMs are used; green boxes are where tools are used.}
    \Description{}
    \label{fig:method}
\end{figure*}

\begin{table*}[t!]
\centering
\caption{List of tools in our current toolbox. More tools may be added to our framework.}
\label{table:tools}
\vspace{-0.5em}
\begin{tabularx}{\textwidth}{llllX}
\toprule
\bf Group & \bf Tool & \bf Input & \bf Output & \bf Description\\
\midrule

\multirow{6}{*}{Lookup} 
                        & \toolname{get\_game\_name} & \multirow{4}{*}{Game ID} & Game name & \small{Return the game name.}\\ \cmidrule{4-5}
                        & \toolname{get\_game\_genre} &  & Game genre & \small{Return the game genre among the 21 predefined categories, e.g., `RPG'.} \\ \cmidrule{4-5}
                        & \toolname{get\_game\_description} &  & Game description &  \small{Return a 2-3 sentence summary of what the game is about and how it is played.} \\ \cmidrule{4-5}
                        & \toolname{get\_game\_rank} &  & Game rank & \small{Return the game rank by number of upvotes.} \\ \cmidrule{3-5}
                        & \toolname{is\_device\_compatible} &  Game ID, Device & True or False & \small{Determine if the game is compatible with the given device, e.g., `CONSOLE'.} \\ 
\midrule
\multirow{2}{*}{Linking}   
                        & \toolname{get\_game\_id\_from\_fuzzy\_name} & Fuzzy name & Game ID & \small{Given an approximate game name, return a game ID that is highly likely to correspond to this game. If no game is found, return nothing. E.g., `MM2' $\rightarrow$ ID for `Murder Mystery 2'} \\ \cmidrule{3-5}
                        & \toolname{fuzzy\_genre\_to\_genres} & Fuzzy genre & Genres & \small{Given a fuzzy genre name, return a list of predefined genres that are likely to correspond to this genre. If no genre is found, return nothing. E.g., `simulation'} $\rightarrow$ [`Simulator/Clicker', `Tycoon/Management Sim'] \\
\midrule
\multirow{5}{*}{Retrieval} 
                        & \toolname{get\_search\_results} & Simple query & \multirow{6}{*}{Game IDs} & \small{Use the search API to return the games relevant to a simple query (maximum 3 words).}\\ \cmidrule{5-5}
                        & \toolname{get\_similar\_games\_cf} & Game ID & 
                        & \small{Use the collaborative filtering API to return `users who played this game also played $\cdots$.'}\\ \cmidrule{5-5}
                        & \toolname{get\_similar\_games\_content} & Game ID &  
                        & \small{Use the SBERT~\cite{reimers2019sentence} embeddings to return the games that have similar descriptions.} \\ \cmidrule{5-5}
                       & \toolname{get\_games\_by\_age\_group} & Age group &  & \small{Get games commonly played among the given age group, e.g., `18-24'.} \\ \cmidrule{5-5}
                       & \toolname{get\_default\_games} & \# games & & \small{Randomly sample games from the top 100 games. May be needed when a user request is too generic.}\\
\midrule
\multirow{2}{*}{Formatting} 
                    & \toolname{get\_game\_info\_str} & Game ID & Formatted info. & \small{Return a string of game information in the following format: `\{game name\} -- \{genre\}. \{description\}'}\\ \cmidrule{3-5}
                    & \toolname{game\_ids\_to\_enum\_game\_info} & Game IDs & Formatted info. & \small{Return a string of enumerated game information in the order of the given list.}\\
\bottomrule
\end{tabularx}
\end{table*}

\section{Methods}

\subsection{Dataset}\label{subsec:data}

\subsubsection{Requests}
We identify a Reddit community /r/Roblox, where users discuss a wide range of topics about Roblox and its games.
Here, we find that some posts are asking for Roblox game recommendations.
We sample the posts by using the Python Reddit API Wrapper (PRAW),\footnote{https://praw.readthedocs.io/en/stable/} using keyphrases such as `recommend me games' and `what games to play'.
We further filter the posts by asking GPT-3.5~\cite{openai2024gpt35} to judge whether the request is asking for game recommendations, and then removing the ones that are not.
This process may still leave a handful of irrelevant posts, such as a game developer asking for recommendations on what game to make.
As such, we manually remove the remaining irrelevant posts (73 out of 629 posts).
We denote the resulting 556 posts as \textit{requests}.

\subsubsection{Human Oracles}
For each request, there are comments from other users in the community that recommend games that are relevant to the request.
We regard these games as \textit{human oracles}. 
When users mention game names, they may not precisely state the exact name, such as referring it with an acronym (e.g., `MM2' instead of `Murder Mystery 2') or dropping out parts of the name (e.g., `Bloxburg' instead of `Welcome to Bloxburg').
To handle this, we ask GPT-3.5 to extract any phrases that might be a game name (to ensure high recall), and link it to real game IDs using the Roblox search API\footnote{Precisely, we use one of our tools, \toolname{get\_game\_id\_from\_fuzzy\_name} (see Table~\ref{table:tools}).} (to ensure high precision).\footnote{To illustrate, `weirdest game on roblox' is a game name; `fun surfing' is not.}
To ensure the quality of oracles, we measure community agreement through the net upvotes of comments. 
For each request, we keep games that have at least one net upvote and discard the rest.
We obtain $553$ requests with at least one oracle.
There are $14.21 (\pm 32.22)$ oracles per request and $2074$ unique games in total.

\subsubsection{From Oracles to Ground-Truth Items}
Human oracles may be noisy (i.e., some games are irrelevant) or insufficient (i.e., there may be more games that are relevant to a request).
We refine the set of recommendations through a two-step process.
First, for each request, we generate a candidate set of games.
This is done by using the oracles: we obtain games similar to the oracle by using two Roblox APIs.\footnote{We use tools: \toolname{get\_similar\_games\_cf} and \toolname{get\_similar\_games\_content} (Table 2).}
Oracles and similar games are added to the candidate set by prioritizing their frequency across all oracles and APIs, with up to $30$ candidates generated per request.
Second, human experts determine whether each candidate is relevant to the request.
These experts are highly knowledgeable about Roblox games, but if they are unfamiliar with a displayed game, they must play it to evaluate its relevance. 
Additionally, to ensure safety, experts are instructed to remove any age-inappropriate games based on the request. 
We denote the resulting games as the \textit{ground-truth items} for a given request.
Due to resource constraints, $208$ requests were processed using the above method.
Each request has an average of $9.06 (\pm 9.03)$ ground-truth games, totaling $1031$ unique games.

\subsection{Proposed Framework: \method}

\subsubsection{System Overview}
Our system overview is depicted in Figure~\ref{fig:method}.
When a user submits a request, an LLM generates a dictionary summarizing the user's preference, denoted as the \textbf{formatted intent}. 
The formatted intent is given as input to the \textbf{tool execution policy}, which selects the tools and arguments to execute and returns an \textbf{execution output} in natural language.
Note that the execution output is not the final recommendation; it contains the relevant information that would augment the LLM with external knowledge (e.g., item information) so that it generates better recommendations.
In the recommendation phase, both the raw request and execution output are provided to the LLM, which generates a list of game names. 
Each game is then linked to a real item in the Roblox database and displayed to the user.

\subsubsection{Formatted Intent Generation}
While it is possible to make LLMs directly generate code policies for tool execution~\cite{liang2023code, wang2024executable}, we later show that this approach is not effective for our task (Section~\ref{sec:experiments}).
Furthermore, from an industry perspective, we want the system to be \textit{transparent} (we can see how the system is operating), and \textit{controllable} (we can easily control and fix how the system works).
In this sense, we propose the following design: let the LLM first process the raw request into a formatted intent \format, and and execute a handcrafted policy \policy\thinspace based on the formatted intent.
This design has several practical benefits: 
(1) it allows us to view the intermediate stage (formatted intent), helping us assess incoming requests and verify whether they are understood or parsed correctly;
(2) instead of depending on LLMs for code generation—which can be a black box and have syntax errors—we rely on human experts for a better understanding and execution of tools;
(3) it yields better performance than using LLM-generated policies.
Specifically, we use the following prompt:

\textit{``Given a user's recommendation request, format the user's preference into a JSON format. 
Fill in the following template of dict[str, dict[str, list]] with the relevant information accurately extracted from the user's request: <template>  <demonstrations>''}.

The <template> consists of preferences and user demographics, where each preference (`like' and `dislike') contains four fields:
\begin{itemize}
    \item Genres: approximate game genres that do not need to match Roblox's official categories exactly
    \item Game names: approximate game names that do not need to match Roblox's game names exactly
    \item Properties: simple keyphrases describing the features or elements of a game
    \item Devices: a subset of `DESKTOP', `PHONE', `TABLET', `CONSOLE', and `VR'.
\end{itemize}

User demographics are composed of two fields:
\begin{itemize}
    \item Ages: age group(s) of user(s) from a subset of `0-8', `9-12', `13-17', `18-24', `25-34', `35plus'.\footnote{We use LLMs to generate devices and age groups from predefined categories since there are only a handful of them and doing so does not require domain knowledge. Entity linking of genres and game names require specialized tools.}
    \item Genders: gender(s) of user(s) inferred from explicit information in the request (e.g., `my son' $\rightarrow$ `MALE').\footnote{After collecting the formatted intents, we notice that genders are rarely mentioned, so we have not created a relevant tool for gender. If a model is deployed within a production platform, incorporating demographics from user account profiles is a feasible enhancement we may consider in future work.}
\end{itemize}

We provide 5 demonstrations, which, compared to no demonstration, results in a more stable generation of formatted intents with the proper template and syntax.

\subsubsection{Toolbox}

Our tools are Python functions, each performing a specific retrieval task that is potentially useful for recommendation.
We present the entire list of tools in Table~\ref{table:tools}, along with each tool's input, output, and description.
Tools are broadly classified into four categories:
\textbf{Lookup} tools return simple game metadata from the Roblox database.
Lookup tools can be used for informing LLMs with item knowledge (e.g., game descriptions), or filtering items based on attributes (e.g., compatible devices).
Although some works propose to employ a single lookup tool by SQL query generation~\cite{wang2024recmind, wang2024what}, this method may not be suitable in many applications, including ours.
For example, the database used in production may not be in a structure where LLMs can generate accurate and efficient SQL queries.
Instead, we propose to have multiple simple tools for accessing the database.
Such design is also important to making the system transparent and controllable. 
\textbf{Linking} tools match game names and genres from user utterances to corresponding entities in the Roblox database. 
These tools are essential for handling real user requests where exact game IDs or genre categories are not used. 
Although implementing a drop-down list in the user interface~\cite{li2018towards} could bypass this issue, we believe it reduces engagement by requiring users to select from a list instead of typing naturally.
\textbf{Retrieval} tools retrieve games that may be relevant to the user's request.
For example, if a user references a game to express their preference, the similarity-search tools retrieve similar games using collaborative filtering (based on similar users) and game content (based on descriptions).
While similar to candidate generators, recommendations are not necessarily confined to the retrieved games.
Instead, the purpose of retrieval tools is to make LLMs be `aware' of the diverse items in the system instead of generating the most popular ones.
Later we show that the absence of these tools results in much lesser diversity of recommended items.
\textbf{Formatting} tools summarize the tool execution results into a natural language format, which would be provided in the prompt for the recommendation stage. 

Note that we do not use ranking tools. 
Instead of using a ranking tool to output the final recommendations~\cite{huang2023recommender},
we let LLMs do the eventual recommendation by having them enumerate a list of items, as we later discuss in Section~\ref{subsubsec:rec}.

\begin{algorithm}[t!]
\caption{Tool execution policy \policy}
\begin{algorithmic}[1]
\State \textbf{Input:} Formatted intent dictionary \format 
\State \textbf{Output:} Results dictionary \augment
\State \textbf{Initialize:} \augment $\gets \{\}$

\For{game in \format[liked games]}
    \State \augment $\gets$ \augment $\cup$ \mygreen{lookup}(game) \Comment{Lookup}
    \State \augment $\gets$ \augment $\cup$ \mygreen{similar}(game) \Comment{Similar}
\EndFor

\State \augment $\gets$ \augment $\cup$ \mygreen{search}(\format[liked genres]) \Comment{Search}

\If{\augment $=\{\}$}
    \State \augment $\gets$ \augment $\cup$ \mygreen{search}(\format[liked properties]) \Comment{Search}
\EndIf

\For{game in \format[disliked games]} 
    \State \augment $\gets$ \augment $\cup$ \mygreen{lookup}(game) \Comment{Lookup}
\EndFor

\State \augment $\gets$ \augment $\cup$ \mygreen{games\_by\_age}(\format[user age groups])  \Comment{Age}

\If{\augment $=\{\}$}
    \State \augment $\gets$ \mygreen{default\_games}(30) \Comment{If the user's request is too generic, i.e., \augment\thinspace is empty so far, randomly sample 30 games from top-100 games.}
\EndIf

\For{game in \augment[similar, search, age results]} \Comment{Filter}
    \If{\mygreen{genre}(game) in \format[disliked genres]}
        \State \augment $\gets$ \augment $\setminus$ game
    \EndIf
    \If{\mygreen{incompatible}(game, \format[preferred devices])}
        \State \augment $\gets$ \augment $\setminus$ game
    \EndIf
\EndFor

\State \augment $\gets$ \mygreen{format}(\augment)  \Comment{Format}

\State \textbf{return} \augment

\end{algorithmic}
\label{algorithm1}
\end{algorithm}

\subsubsection{Tool Execution}
We describe our tool execution policy \policy: \format$\rightarrow$\augment\thinspace in Algorithm~\ref{algorithm1}.\footnote{For presentation simplicity, we omit linking tools and abbreviate tool names.}
The policy goes through each (key, value) in the formatted intent and runs the corresponding tools, adding information to \augment\thinspace that would potentially be helpful to the recommendation stage.
Then the policy goes through \augment\thinspace again and filters items that can be sources of noise (e.g., games that are incompatible with the user's preferred devices).
While we can skip the filtering and let the LLM disregard irrelevant items in the final recommendation stage, we find that simply filtering items in advance improves recommendation performance.
Finally, the policy uses the formatting tools to convert \augment\thinspace into a readable format to be passed into the recommendation stage.
For example, 
\begin{quote}
    \{`Users who played id0 also played': [id1, id2, $\cdots$]\}
\end{quote}
 
 becomes 

\begin{quote}
    Users who played `Da Amazing Bunker Simulator' also played:
    \begin{enumerate}[label=\arabic*.]
        \item RetroStudio --- Genre: Sandbox. This game allows players to create $\cdots$
    \end{enumerate}
\end{quote}

\subsubsection{Recommendation}\label{subsubsec:rec}

To generate high-quality recommendations, a model needs to accurately understand complex and nuanced requests.
LLMs excel in natural language understanding to such an extent that they surpass traditional, smaller models at conversational recommendation~\cite{he2023large}.
As such, instead of having a separate tool (e.g., for ranking) to generate the final recommendations, we prompt an LLM with the raw request, tool execution output \augment, and an instruction to generate a list of relevant items.
This method utilizes the LLM's language capability (i.e., understanding raw request) and augments its weakness by providing external knowledge (i.e., tool execution output).
We use the following instruction: 
\textit{`Given the following request, provide recommendations. Enumerate 20 Roblox game names (1., 2., ...) in the order of relevance. Don't say anything else.'}
We augment the LLM with \augment\thinspace by adding:
\textit{`Using the above information along with your own knowledge and reasoning, provide the best recommendations that fulfill the request.'}

\begin{table*}[t!]
\centering
\begin{tabularx}{\textwidth}{c|l|c|ccc|cc|cc}
\toprule
 & \multirow{2}{*}{\bf Method} & \multicolumn{1}{c|}{\bf Factuality} & \multicolumn{3}{c|}{\bf Relevance}  & \multicolumn{2}{c|}{\bf Novelty} & \multicolumn{2}{c}{\bf Coverage} \\ 
\cmidrule{3-10}
&                                  & Factual (↑) & Hit (↑)      & Precision (↑)   & Sim (↑)      & Pop50 (↓)     & RPop50 (↓)    & Entropy (↑)   & MaxFreq (↓) \\ 
\cmidrule{2-10}
& Pop                              & 1.00 \soften{1.00} & .08 \soften{.14} & .02\soften{.04}  & .91 \soften{.89} & 1.00 \soften{1.00} & 10.31 \soften{7.97}  & 5.61 \soften{5.64}  & 0.15 \soften{.12}\\
\midrule
\multirow{4}{*}{LLaMA-405B} 
& Base LLM                         & .84 \soften{.88} & \underline{.22} \soften{\bf.23} & \underline{.06} \soften{\underline{.06}}  & .91 \soften{.88} & .35 \soften{.48} & 3.60 \soften{3.84}  & 7.16 \soften{6.57}  &  .36 \soften{.53}\\
& Base LLM + Div                   & .68 \soften{.70} & .13 \soften{.16} & .03 \soften{.04}  & .86 \soften{.84} & .15 \soften{\bf.17} & 1.52 \soften{\underline{1.39}}  & 7.66 \soften{7.63}  &  .18 \soften{.27}\\
& \method w/ \policyllm    & \underline{.98} \soften{\underline{.98}} & \underline{.22} \soften{18} & .05 \soften{.05}  & \underline{.92} \soften{\bf.89} & \underline{.14} \soften{\bf.17} & \underline{1.39} \soften{\bf1.35}  & \textbf{8.97} \soften{\bf9.18}  &  \underline{.10} \soften{\bf.12}\\
& \bf \method w/ \policy            & \textbf{1.00} \soften{\bf.99} & \textbf{.25} \soften{\bf.23} & \textbf{\bf.07} \soften{\bf.07}  & \textbf{.93} \soften{\bf.89} & \textbf{.13} \soften{.21} & \textbf{1.38} \soften{1.63}  & \underline{8.81} \soften{\underline{8.85}}  &  \textbf{.05} \soften{\underline{.16}}\\

\midrule
\multirow{4}{*}{GPT-4o} 
& Base LLM                         & .90 \soften{.94} & \underline{.26} \soften{\bf.29} & \underline{.07} \soften{\bf.09}  & .90 \soften{.88} & .42 \soften{.56} & 4.34 \soften{4.48}  & 7.17 \soften{6.64}  &  .20 \soften{.39}\\
& Base LLM + Div                   & .59 \soften{.64} & .16 \soften{.18} & .04 \soften{.05}  & .73 \soften{.73} & \textbf{.11} \soften{\bf.12} & \textbf{1.10} \soften{\bf.96}  & 8.15 \soften{8.53}  &  \textbf{.07} \soften{\bf.10}\\
& \method w/ \policyllm    & \underline{.98} \soften{\bf.99} & .22 \soften{.19} & .06 \soften{.06}  & \textbf{.93} \soften{\bf.90} & \underline{.16} \soften{\underline{.21}} & \underline{1.60} \soften{\underline{1.67}}  & \textbf{8.73} \soften{\bf8.97}  &  .10 \soften{\bf.10}\\
& \bf \method w/ \policy            & \textbf{.99} \soften{\bf.99} & \textbf{.27} \soften{\underline{.24}} & \textbf{.08} \soften{\underline{.08}}  & \textbf{.93} \soften{\underline{.89}} & .17 \soften{.27} & 1.71 \soften{2.14}  & \underline{8.68} \soften{\underline{8.71}}  &  \textbf{.07} \soften{\underline{.12}}\\
\bottomrule
\end{tabularx}
\label{tab:res5}
\end{table*}

\begin{table*}[t!]
\centering
\begin{tabularx}{\textwidth}{c|l|c|ccc|cc|cc}
\toprule
 & \multirow{2}{*}{\bf Method} & \multicolumn{1}{c|}{\bf Factuality} & \multicolumn{3}{c|}{\bf Relevance}  & \multicolumn{2}{c|}{\bf Novelty} & \multicolumn{2}{c}{\bf Coverage} \\ 
\cmidrule{3-10}
&                                   & Factual (↑) & Hit (↑)      & Precise (↑)   & Sim (↑)      & Pop50 (↓)     & RPop50 (↓)    & Entropy (↑)   & MaxFreq (↓) \\ 
\cmidrule{2-10}
& Pop                               & 1.00 \soften{1.00} & .15 \soften{.19}      & .02 \soften{.03}      & .93 \soften{.90}      & 1.00 \soften{1.00}      & 11.23 \soften{8.40}      & 5.63 \soften{5.64}      & .26 \soften{.24}      \\
\midrule
\multirow{4}{*}{LLaMA-405B} 
& Base LLM                          & .79 \soften{.83}     & .25 \soften{\underline{.28}}      & \underline{.04} \soften{\underline{.05}}      & .92 \soften{.89}   & .28 \soften{.40}      &  3.19 \soften{3.32}      & 7.68 \soften{7.26}      & .43 \soften{.60}      \\
& Base LLM + Div                    & .64 \soften{.67}      & .16 \soften{.19}    & .02 \soften{.03}       & .87 \soften{.85}       & \textbf{.11} \soften{\underline{.14}}   & \textbf{1.25} \soften{\textbf{.21}}      & 7.82 \soften{7.79}      & .17 \soften{.30}      \\
& \method w/ \policyllm       & \underline{.98} \soften{\underline{.98}}  & \underline{.30} \soften{.24}   & \underline{.04} \soften{.04}  & \underline{.93} \soften{\underline{.90}}     &  .13 \soften{\textbf{.17}}  & 1.50 \soften{\underline{1.43}}   & \textbf{9.50} \soften{\textbf{9.60}}   & \underline{.16} \soften{\underline{.23}}      \\
& \bf \method w/ \policy  & \textbf{1.00} \soften{\textbf{.99}}   & \textbf{.36} \soften{\textbf{.31}}   & \textbf{.05} \soften{\textbf{.06}}  & \textbf{.94} \soften{\textbf{.91}}   & \underline{.12} \soften{.19}  & \underline{1.31} \soften{1.63}  & \underline{9.48} \soften{\underline{9.43}}     &  \textbf{.10} \soften{\textbf{.19}}      \\
\midrule
\multirow{4}{*}{GPT-4o} 
& Base LLM                          & .89\soften{.93}   & \underline{.36} \soften{\textbf{.38}}   & \textbf{.06} \soften{\textbf{.08}}    & .92 \soften{.89}      & .38 \soften{.53}   & 4.24 \soften{4.43}    & 7.62 \soften{7.08}    & .28 \soften{.50}     \\
& Base LLM + Div                    & .57\soften{.61}   & .20 \soften{.23}   & .03 \soften{.04}   & .74 \soften{.74}    & \textbf{.09} \soften{\textbf{.12}}   & \textbf{1.04} \soften{\textbf{.97}}    & 8.37 \soften{8.63}    & \textbf{.08} \soften{\textbf{.14}}      \\
& \method w/ \policyllm       & \underline{.98} \soften{\textbf{.99}}   & .33 \soften{.28}   & .05 \soften{.05}     & \textbf{.94} \soften{\textbf{.91}}      & .15 \soften{\underline{.22}}    & 1.72 \soften{\underline{1.84}}    & \underline{9.23} \soften{\textbf{9.33}}    & .13 \soften{\underline{.15}}      \\
& \bf \method w/ \policy   & \textbf{.99}\soften{\textbf{.99}}   & \textbf{.38} \soften{\underline{.33}}   & \textbf{.06} \soften{\underline{.06}}   & \textbf{.94} \soften{\textbf{.91}}   & \underline{.14} \soften{.25}   & \underline{1.61} \soften{2.13}   & \textbf{9.31} \soften{\underline{9.21}}    & \underline{.12} \soften{.24}   \\
\bottomrule
\end{tabularx}
\vspace{0.5em}
\caption{Results for top 5 (above) and 10 (below) recommendations, on human-annotated and full (colored \soften{grey}) datasets.}
\label{tab:res}
\vspace{-1em}
\end{table*}

\section{Experiments}\label{sec:experiments}

\subsection{Evaluation Metrics}
We use multiple evaluation metrics for relevance, novelty, and coverage of recommended items.
Additionally, since we are using LLMs as recommenders, we measure factuality to understand whether models are hallucinating.

\subsubsection{Relevance}
\textbf{Hit@k} evaluates whether a ground-truth item is included in the top-k recommendations. 
\textbf{Precision@k} is the proportion of ground-truth items in the recommendations.
\textbf{Similar@k} is the similarity of ground-truth items and recommended items by computing the cosine distance between the embedding centroids. 
In our work, we use SimCSE~\cite{gao2021simcse} embeddings obtained from item descriptions.
We average each of the metrics across all requests.

\subsubsection{Novelty}
The concept of novelty in recommender systems can vary, but it is commonly linked to an item's popularity, such as the number of ratings it has received~\cite{vargas2011novelty, kaminskas2016novelty}.
In a similar vein, we use a metric that uses item popularity, where \textbf{Pop50@k} is the proportion of items in the top 50 most popular (or well-known) games, ranked by upvotes. 
Lower values are better since popular items are often listed on the Roblox front page, and our objective is to help users discover unfamiliar items.
We also use \textbf{RPop50@k}, which computes the ratio of Pop50@k for the recommended items to that of the ground-truth items.
Closer value to 1 indicates that the recommendations are as novel as the ground-truth items.

\subsubsection{Coverage}
\textbf{Entropy@k} measures the diversity of recommended items across all requests, formally computed by the following equation: $\text{Entropy@k} = - \sum_{i} p_i \log(p_i)$, where \( p_i \) is defined as the frequency of item \( i \) across the top-k recommendations. 
Higher entropy indicates a wider coverage of items~\cite{kaminskas2016novelty, qin2013promoting}.
\textbf{MaxFreq@k} identifies the most frequently recommended item, and computes the proportion of requests that this item is recommended. 
For example, if `Adopt Me!' appears in the top-10 list in 60\% of the requests, then MaxFreq@10 is 0.60.
A lower value is preferable, as it indicates that the system avoids recommending the same item repeatedly.

\subsubsection{Factuality}
\textbf{Factual@k} measures the proportion of real items in the top-k list.
If the tool \toolname{get\_id\_from\_fuzzy\_name} returns nothing, we regard the game name as hallucinated.
While factuality can be easily addressed by displaying only the actual items to the user, it remains an important metric for understanding model performance.
We compute other metrics after filtering out hallucinated items.

\begin{figure*}[t]
    \centering
    \begin{subfigure}[t]{0.48\linewidth} 
        \centering
        \includegraphics[width=\linewidth]{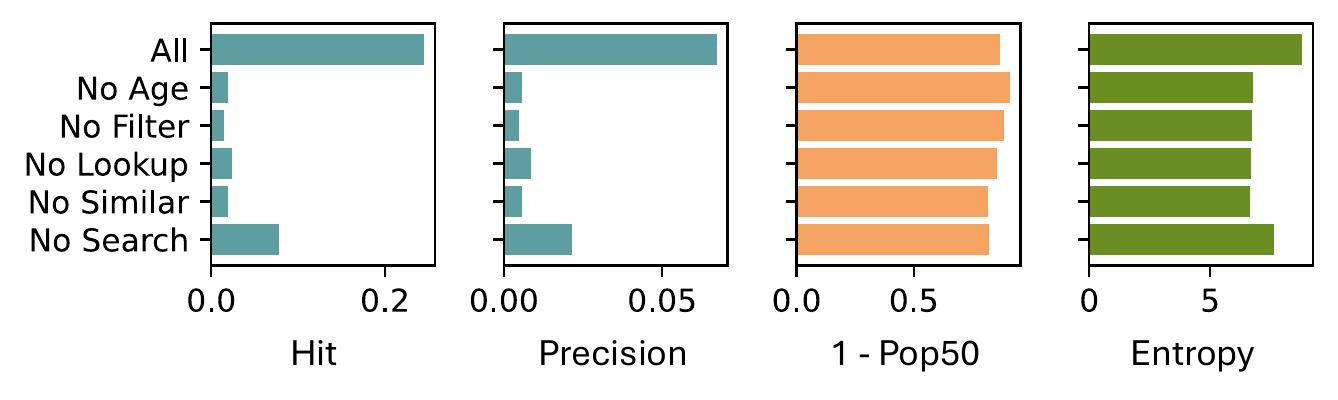}
    \end{subfigure}
    \begin{subfigure}[t]{0.48\linewidth} 
        \centering
        \includegraphics[width=\linewidth]{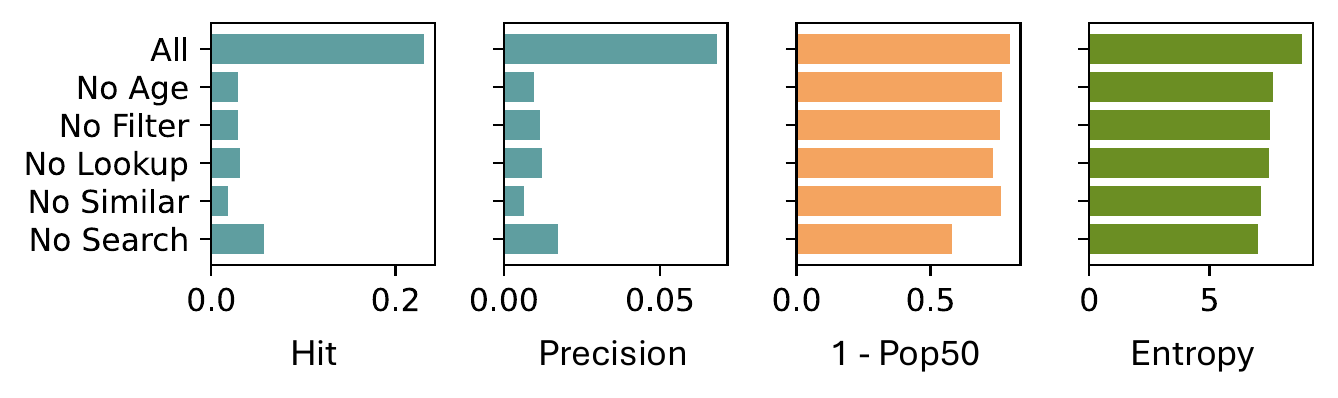}
    \end{subfigure}
    \hfill    
    \begin{subfigure}[t]{0.48\linewidth} 
        \centering
        \includegraphics[width=\linewidth]{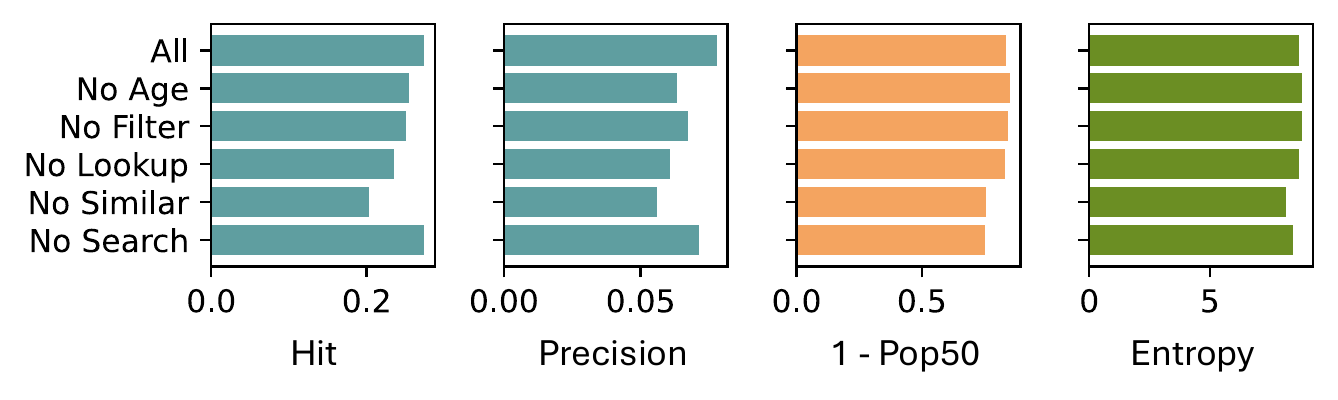}
        \caption{Annotated dataset: LLaMA-405B (above) GPT-4o (below)}
    \end{subfigure}
    \begin{subfigure}[t]{0.48\linewidth} 
        \centering
        \includegraphics[width=\linewidth]{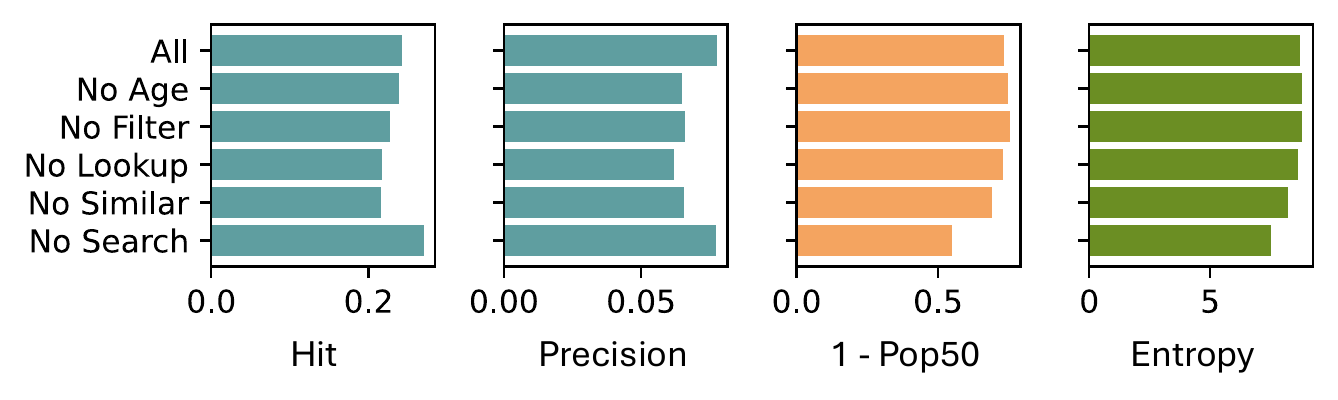}
        \caption{Full dataset: LLaMA-405B (above) GPT-4o (below)}
    \end{subfigure}

    \caption{Ablation study. Recommendations are more relevant if we use more tools. 
    One exception is when removing the search tool for GPT-4o: relevance increases, but this comes at a relatively large cost to both novelty (1$-$Pop50) and diversity (Entropy).
    Above are results for $k=5$ and we observe similar trends for different $k$ values.}
    \Description{}
    \label{fig:ablation}
\end{figure*}
 
\subsection{Setup}

\subsubsection{LLMs}
We use LLaMA-405B~\cite{meta2024llama} and GPT-4o~\cite{openai2024gpt4}.
The temperatures of LLMs are set to $0$ for deterministic results.\footnote{We also tried other temperature values, but the differences were insignificant.} 
For simplicity, we use the same LLMs for formatting and recommendation. 
In practice, the two stages can be run by different LLMs. 

\subsubsection{Baselines}
Previous works (see Section~\ref{sec:related}) use zero-shot~\cite{he2023large, sanner2023large, yoon2024imagery} or tool-augmented~\cite{huang2023recommender, kemper2024retrieval, li2024incorporating, wang2024recmind, xi2024memocrs} LLMs for CRS. 
Since the tools in each paper are often domain-specific and difficult to apply in our work, we perform ablation tests to show the necessity of a large number of tools, which distinguishes us from existing methods.
RAG-based approaches that retrieve similar queries from the training corpus ~\cite{xie2024neighborhood} are also unsuitable for our setting due to a small volume of available queries (which we entirely use for evaluation). 
While some works fine-tune LLMs with traditional user-item data or synthetic queries~\cite{kang2023llms, zhang2023recommendation}, we do not consider them as baselines since we want to incorporate external knowledge without the cost of fine-tuning.
As such, our baselines are as follows:

\begin{itemize}
    \item \textbf{Pop} randomly selects \( k \) items from the top-50 list.
    \item \textbf{Base LLM} is an LLM without any tool augmentations.
    \item \textbf{Base LLM + Div} is a slight variant that encourages the base LLM to generate lesser-known items by simply adding the following instruction: \textit{`The games should be diverse and not too well-known (should be new to the user).'} 
    \item \textbf{\method w/ \policyllm} replaces the handcrafted policy \policy\thinspace with LLM-generated ones, to observe whether LLMs can generate better policies than \policy, as previous works suggest~\cite{liang2023code, huang2023recommender}.
    We provide the LLM with the raw request, formatted intent, and a list of available tools and instruct to generate a code in Python that outputs \augment.\footnote{We also tried providing a demonstration of a handcrafted policy, but we observe that this makes LLM replicate the handcrafted policy.}
\end{itemize}

\subsection{Results}

We organize the results into multiple research questions.
Results for Q1-3 are in Table~\ref{tab:res}, and the ablation study for Q4 is in Figure~\ref{fig:ablation}.

\boldheading{Q1. Is \method more effective than base LLMs?}
\method outperforms base LLMs in all metrics for the human-annotated dataset (see Table~\ref{tab:res}). 
For the full dataset, \method outperforms base LLaMA-405B in all metrics and GPT-4o in all but Hit and Precision;
this discrepancy could be attributed to the lack of accurate ground-truth items for the full dataset.
Base LLMs have particularly poor novelty and coverage; LLaMA-405B recommends top-50 items $\times3.19$ more frequently than the ground-truths, and recommends the most frequent item (`Natural Disaster Survival') in 43\% of requests.
This is in contrast to \method, where LLaMA-405B recommends top-50 items only $\times1.31$ more than the ground-truths, and recommends the most frequent item in 10\% of requests.
While \method achieves near-perfect factuality (> 99\%), 
base LLMs generate hallucinations among 21\% (LLaMA-405B) and 11\% (GPT-4o) of top-10 recommendations.

\boldheading{Q2. Is fixed \policy\thinspace better than LLM-generated policies?}
We experiment to see if LLMs can generate their own policies, \policyllm, per request using the same toolbox, to determine if they can create more effective, customized policies.
We find that although LLMs generate reasonable policies, relevance metrics significantly drop compared to simply using the fixed policy \policy.
We observe high coverage (Entropy) in some cases, but the overall results show that there is little or no advantage using \policyllm\thinspace over fixed \policy, especially considering that the former approach is less transparent and controllable.
That said, \method with \policyllm\thinspace consistently outperforms base LLMs in factuality, novelty, and coverage, and occasionally in relevance, suggesting that retrieving any relevant results is preferable to none for factual and diverse recommendations.

\boldheading{Q3. Can we prompt LLMs to recommend more diverse items?}
Base LLMs indeed generate more diverse items (higher novelty and coverage) when explicitly prompted to do so, but this leads to a significant loss in relevance (55-64\% of unprompted) and factuality (64-81\% of unprompted).
While simple prompting yields even higher novelty than \method when $k=10$, the differences are relatively small (1.25 v.s. 1.31 for LLaMA and 1.04 vs.~1.61 for GPT-4o in RPop).

\boldheading{Q4. Do we need all the tools?}
Figure~\ref{fig:ablation} shows the results of our ablation study, where we remove each tool to observe the impact on performance. 
We find that using all the tools generally improves relevance, with two unexpected results.
One is that the performance of LLaMA-405o significantly drops when any tool is omitted.
A possible explanation is that augmenting with partial information may mislead the model (e.g., by providing similar games but not age-relevant games).
The model may also be sensitive to noise when the filtering tool is not used.
Another interesting result is that dropping the search tool can slightly increase relevance (although at the notable cost of novelty and converage) for GPT-4o.
To understand this, we examined the search tool’s outputs. 
One issue is that the Roblox search API sometimes returns noisy results, such as retrieving low-quality games. 
But a more fundamental problem is that many user-described properties, such as `sweet', `not too horror', `no progression', `nice people', and `unique premise', can be ambiguous or incompatible with search queries.
\method is intended to handle such nuanced requests by letting LLMs understand the request holistically (e.g., `sweet' as the game `Oobja', or `not too horror' as less intense than `The Mimic') and use the provided game descriptions to match them with the request.
However, their descriptions alone may not provide enough context to accurately match games with requests.\footnote{For example, in the case of `The Mimic,' the description mentions it is a horror game with jumpscares, but it does not convey the intensity of the horror compared to other games. Such information could be gleaned from user opinion data, such as reviews.}
One way to address this issue is to obtain descriptions of actual gameplay or user opinion, which we consider as future improvements. 
In terms of novelty and coverage, using all tools yields similar or better results than omitting any.

\section{Deployment}

\begin{figure}[t!]
    \centering
    \includegraphics[width=0.95\linewidth]{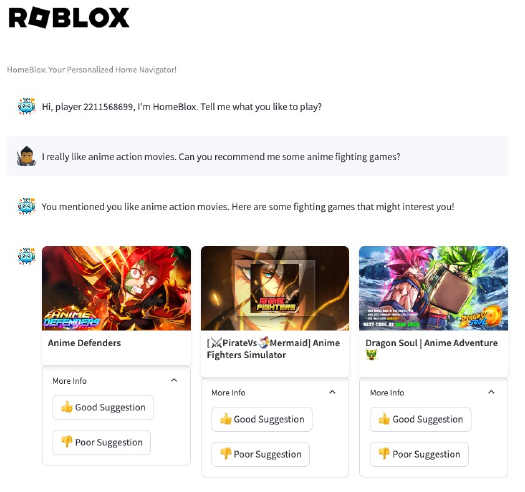}
    \caption{Screenshot of the deployed UI. We add simple greeting and explanations for a more natural conversation, and thumbs up and down buttons for obtaining feedback.}
    \Description{}
    \label{fig:demo}
\end{figure}

To perform a feasibility study and identify the best implementation practices, we launch an internally hosted chatbot (see Figure~\ref{fig:demo}).
Our application is built on a full-stack server using Streamlit~\cite{streamlit}, which simplifies creating an interactive UI and managing backend operations.
We deploy the application in an internal datacenter using HashiCorp's Nomad and Consul~\cite{hashicorp} for cluster orchestration, deployment, and configuration. 
Several key areas are under evaluation to assess the feasibility of transitioning the chatbot to production.
First is ensuring system safety by preventing irrelevant queries, policy violations, and jailbreak attempts (see Section~\ref{sec:ethical}).
Second is latency and scalability. Our current chatbot takes several seconds per query to generate results. Further studies are necessary to understand user tolerance for latency and explore techniques to enhance inference efficiency at scale.
\section{Related Work}\label{sec:related}

\boldheading{Conversational Recommender Systems.}
There are two categories of works based on the evaluation approach: interactive and dataset-based.
In interactive evaluation, a user simulator replaces real users, and the problem is often framed into item or attribute selection for preference elicitation~\cite{christakopoulou2016towards, sun2018conversational, lei2020estimation, zhang2020conversational}.
In this approach, user simulators may fall short of reflecting real users~\cite{yoon2024evaluating}.
Dataset-based evaluation recommends items given prior utterance. 
Most existing datasets are crowd-sourced~\cite{li2018towards, moon2019opendialkg, hayati2020inspired}, where workers role-play as seeker and recommender.
Early works propose using two separate modules, language understanding and recommendation, while more recent work suggests merging the two~\cite{wang2022towards}.
Most recently, zero-shot LLMs have shown to outperform all previous methods, especially for complex user utterances~\cite{he2023large}.

\boldheading{LLMs for Recommendation.}
Since LLMs take language inputs, most recommendation task that uses an LLM inherently becomes `conversational'.
Often, queries are generated by inserting non-CRS datasets (e.g., user-item interactions) into templates~\cite{geng2022recommendation, harte2023leveraging, hou2024large, kang2023llms, li2023prompt, wang2024rdrec}.
LLMs can be fine-tuned with such queries~\cite{mysore2023large, zhang2023recommendation}, and further enhanced by incorporating collaborative filtering information during training~\cite{kim2024large, yang2024item, zheng2023adapting, zhu2024collaborative}.
In contrast, our focus is on the user requests expressed in their own words, not bound in templates.

\boldheading{Tool-Augmented LLMs.}
Recent works explore using LLMs to create agents that can perform complex interactive tasks.
Applications include robotic control~\cite{brohan2023saycan}, scientific reasoning~\cite{lin2024swiftsage}, and question answering~\cite{shinn2024reflexion, yao2022react}.
Solving such tasks often requires using tools~\cite{li2023apibank, qin2023toolllm}. 
Tools are functions external to the LLM~\cite{wang2024what}, and can help agents access external knowledge bases~\cite{hao2024toolkengpt, yao2022react}, perform arithmetic operations~\cite{schick2024toolformer, hao2024toolkengpt}, use specialized models~\cite{shen2024hugginggpt, lu2024chameleon}, and interact with the world~\cite{wang2023voyager, zheng2024natural}.
Agents can even create simple tools and add them to the toolbox~\cite{wang2024trove, yuan2023craft}.
Some works explore the possibility of having the agent generate a policy for using tools~\cite{liang2023code, wang2024executable}, but we have shown in our experiments that using a fixed policy is more effective for our task.

\boldheading{Tool-Augmented LLMs for Recommendation.}
In recommendation, access to external knowledge is crucial because items are frequently added or removed, their information is updated (e.g., content updates or shifts in popularity), and external factors (e.g., seasonal demand) can influence user preference.
As such, recent works propose tool-augmented LLMs for recommendation~\cite{huang2023recommender, kemper2024retrieval, li2024incorporating, wang2024recmind, xi2024memocrs} to retrieve relevant information from an external knowledge base.
From a practical perspective, we are faced with several limitations in directly applying this work:
queries are often synthetic~\cite{huang2023recommender, kemper2024retrieval, wang2024recmind}, which are different from real users; 
tools used in previous work are often unavailable in some use cases including ours, e.g., review-based item retrieval~\cite{kemper2024retrieval}.
While demonstration papers~\cite{friedman2023leveraging, kemper2024retrieval, li2024incorporating} focus on implementing working systems, our work complements these efforts by providing extensive evaluation.

\section{Conclusion}

This work aims to advance practical conversational recommender systems by collecting a dataset of real user requests and proposing a novel approach to augmenting large language models with multiple tools. 
Our study includes comprehensive experiments and deployment insights. 
One limitation is we focus on game recommendations, which may not generalize to other domains. 
Additionally, the Reddit dataset may not fully represent all user types. 
As future work, we plan to develop models based on larger datasets.

\section{Ethical Considerations}
\label{sec:ethical}

In developing our system, we prioritize ethical considerations, particularly in the areas of fairness, diversity, and system integrity.
To address fairness and diversity, the evaluation of our system is designed with native support for these principles, using carefully processed datasets and beyond-accuracy metrics to ensure equitable recommendations across users and items. 
In terms of integrity, we implement dedicated modules to handle the following:

\begin{itemize}
\item{Jailbreak prevention:}
A mechanism to protect against external manipulation and unauthorized system exploitation.

\item{Integrity verification:}
A mechanism to ensure the safety of recommendations and the words used by the conversational system, ensuring reliable outputs.
\end{itemize}

While these measures significantly reduce the risks associated with fairness, diversity, and integrity, it is important to acknowledge that due to the inherent complexity of large language models, these issues cannot be entirely eliminated. 
The field is rapidly evolving, and ongoing research is essential to further refine and enhance these protections.
In summary, our system incorporates robust solutions to address ethical concerns, though we recognize the need for continuous improvement as part of the broader research landscape.

\bibliographystyle{ACM-Reference-Format}
\bibliography{main}


\end{document}